\pdfoutput=1

\documentclass[11pt]{article}

\usepackage[final]{acl}

\usepackage{times}
\usepackage{latexsym}
\usepackage{listings}
\usepackage{xcolor} 
\usepackage[T1]{fontenc}

\usepackage[utf8]{inputenc}

\usepackage{microtype}
\usepackage{amsmath}
\usepackage{inconsolata}
\usepackage{multirow}
\usepackage{subcaption}

\usepackage{graphicx}

\title{DiagramEval: Evaluating LLM-Generated Diagrams via Graphs}

\author{
 \textbf{Chumeng Liang\textsuperscript{1}},
 \textbf{Jiaxuan You\textsuperscript{1}}
\\
 \textsuperscript{1}University of Illinois Urbana-Champaign
\\
 \small{
   \href{mailto:chumengl@illinois.edu}{chumengl@illinois.edu}, \href{mailto:jiaxuan@illinois.edu}{jiaxuan@illinois.edu}
 }
}

\begin{document}
\maketitle
\begin{abstract}
Diagrams play a central role in research papers for conveying ideas, yet they are often notoriously complex and labor-intensive to create. Although diagrams are presented as images, standard image generative models struggle to produce clear diagrams with well-defined structure. We argue that a promising direction is to generate demonstration diagrams directly in textual form as SVGs, which can leverage recent advances in large language models (LLMs).
However, due to the complexity of components and the multimodal nature of diagrams, sufficiently discriminative and explainable metrics for evaluating the quality of LLM-generated diagrams remain lacking. In this paper, we propose DiagramEval, a novel evaluation metric designed to assess demonstration diagrams generated by LLMs. Specifically, DiagramEval conceptualizes diagrams as graphs, treating text elements as nodes and their connections as directed edges, and evaluates diagram quality using two new groups of metrics: node alignment and path alignment. For the first time, we effectively evaluate diagrams produced by state-of-the-art LLMs on recent research literature, quantitatively demonstrating the validity of our metrics. Furthermore, we show how the enhanced explainability of our proposed metrics offers valuable insights into the characteristics of LLM-generated diagrams. Code: \url{https://github.com/ulab-uiuc/diagram-eval}.
\end{abstract}

\section{Introduction}

Diagrams play a central role in research papers for conveying ideas, \textit{e.g.}, the diagram of Transformer \cite{vaswani2017attention} has played a pivotal role in presenting and publicizing the idea, making it one of the most cited deep learning papers. However, generating high-quality diagrams is often notoriously complex and labor-intensive to create.
Consequently, automated diagram generation is a central challenge in AI-assisted scientific discovery~\citep{eger2025transforming}, potentially saving millions of hours for researchers while improving the quality of research publications. Although diagrams are presented as images, standard image generative models struggle to produce clear diagrams with well-defined structure \citep{zala2023diagrammergpt}. We argue that a promising direction is to generate demonstration diagrams directly in textual form as SVGs, which can leverage recent advances in large language models (LLMs). 

Existing methods on automated diagram generation with LLMs rely heavily on proprietary LLMs, either through direct planning~\citep{mondal2024scidoc2diagrammer,zhang2024scimage,cui2025draw} or as assistance in diagram generation~\citep{belouadi2023automatikz,belouadi2024detikzify,cui2025draw}. Given this reliance, advancements in cutting-edge proprietary LLMs, such as Claude 3.7 Sonnet~\citep{anthropic2025claude} and Gemini 2.5 Pro~\citep{deepmind2025gemini}, directly enhance automated diagram generation through SVG code generation capabilities~\citep{blecher2023nougat}.

However, existing evaluation metrics lack the expressiveness needed to differentiate diagrams of varying quality. Most benchmarks employ model-based, diagram-level metrics~\citep{hessel2021clipscore,fu2023dreamsim}, which were originally designed for general text-to-image tasks rather than text-within-image generation specific to diagrams~\citep{rodriguez2023ocr}. These metrics evaluate entire diagrams using general vision models, which limits their explainability and their ability to accurately capture detailed logical correctness, resulting in only moderate correlation with human judgments~\citep{eger2025transforming}. Thus, there is a critical need for new metrics.

\begin{figure*}[t]
  \includegraphics[width=0.95\textwidth]{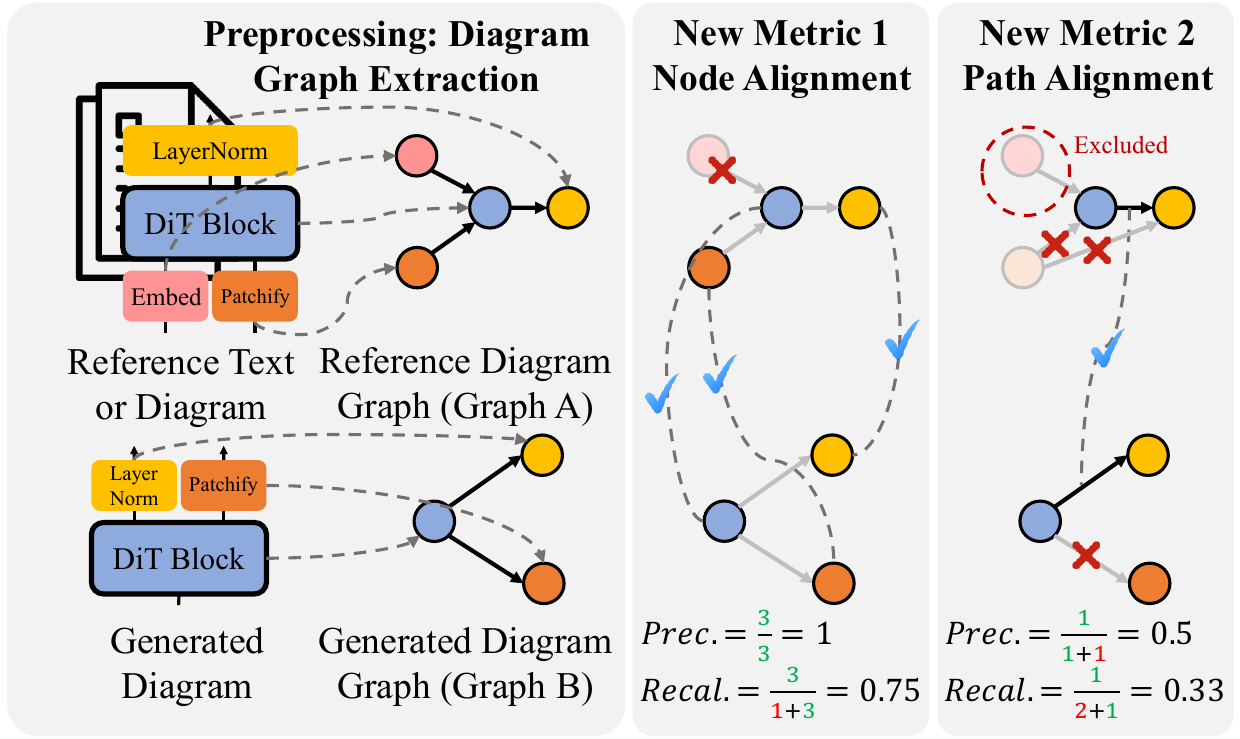}
  \caption{\textbf{DiagramEval framework overview}. By considering both research paper context and paper diagrams as directed graphs, we use information precision and recall among nodes and edges of the generated diagram graph and those of the reference graph (from reference diagrams or paper context) to measure the generation quality of paper diagrams. }
  \label{fig:1}
\end{figure*}

In this paper, we propose DiagramEval, a novel evaluation metric designed to assess demonstration diagrams generated by LLMs. As demonstrated in Figure~\ref{fig:1}, DiagramEval conceptualizes diagrams as graphs, treating text elements as nodes and their connections as directed edges, and evaluates diagram quality using two new groups of metrics: node alignment and path alignment. Specifically, we transform diagrams into SVG format, parsing text elements as nodes. Connections between elements are then extracted as edges. Our metrics quantify the node and path (multi-hop edges) alignment between generated and ground-truth diagrams using precision, recall, and F1 scores. For the first time, we effectively evaluate diagrams produced by state-of-the-art LLMs on recent research literature, validating the proper consistency between our metrics and existing metrics. We also demonstrate metric statistics and case studies that reveal some previously unrecognized shortcomings of existing metrics due to their moderate explainability, for example, over-sensitivity to spatial layouts and other visual elements and metric hacking. We furthermore show empirically how the enhanced explainability of our metrics helps overcome these shortcomings and meanwhile offers valuable insights into the characteristics of LLM-generated diagrams. 

\textit{Based on our proposed metrics, we build a benchmark to evaluate the quality of LLM-generated paper diagrams. We implement an automated pipeline to extract context and diagrams from preprint versions of research papers on Arxiv only with their titles. With this highly available data source, we construct a simple agent for diagram generation with Gemini 2.5 Flash Image, the proprietary image generative models from Google. Our metrics are then used to evaluate the quality of generated diagrams. \textbf{To the best of our knowledge, this is the first benchmark specified for evaluating LLM-generated paper diagrams.} Code: \url{https://github.com/ulab-uiuc/diagram-eval}.}

\section{Related Works} 

Extensive research has addressed both the understanding~\citep{han2023chartllama,liu2023mmc,wang2024charxiv,hu2024mplug} and generation~\citep{maddigan2023chat2vis,yang2024chartmimic} of scientific images, with specific efforts targeting diagram generation~\citep{zala2023diagrammergpt,zhang2024scimage,mondal2024scidoc2diagrammer,cui2025draw}. Current automated evaluation metrics fall into two main categories: \textbf{1)} diagram-to-diagram similarity~\citep{fu2023dreamsim}, and \textbf{2)} caption-to-diagram similarity~\citep{zhang2019bertscore,hessel2021clipscore}. Both metric types primarily measure overall similarity between generated diagrams and references in latent space, neglecting detailed logical accuracy, such as verifying element-wise connections against the paper context. Closely related to our approach is the VPEval metric~\citep{cho2023visual}, which evaluates element and connection accuracy via visual question answering (VQA). However, VPEval often struggles to accurately identify connections even in diagrams involving common knowledge, requiring manual annotations~\citep{zala2023diagrammergpt}. Our approach extends the concept of VPEval by leveraging vision-language models (VLMs) and file parsing to reliably extract elements and connections, thus better suited for evaluating diagrams generated from academic papers.

\section{Preliminary}
\label{sec:bg}
\paragraph{Problem Definition}

Following existing research working on diagram generation~\citep{zala2023diagrammergpt,mondal2024scidoc2diagrammer}, we define the task of automated generation of scientific
diagrams from academic papers: Given the input corpus
$\{T, c, \pi\}$, which includes original paper context $T$,
original diagram captions $c_{orig}$, and layout captions $c_{layout}$, our goal is to \textbf{generate a diagram $\mathcal{D}$ which demonstrates the overall idea of research paper $T$}. We have a groundtruth diagram $\mathcal{D}_{gt}$, which we use to generated the layout captions with an independent vision language model. $\mathcal{D}_{gt}$ also serve as reference to evaluate generated diagrams. As mentioned, the diagram generation is done in the SVG format~\citep{blecher2023nougat}, which is both editable and widely adopted by proprietary LLMs.

\section{Methodology}
\label{sec:met}
\begin{figure*}[t]
  \includegraphics[width=\textwidth]{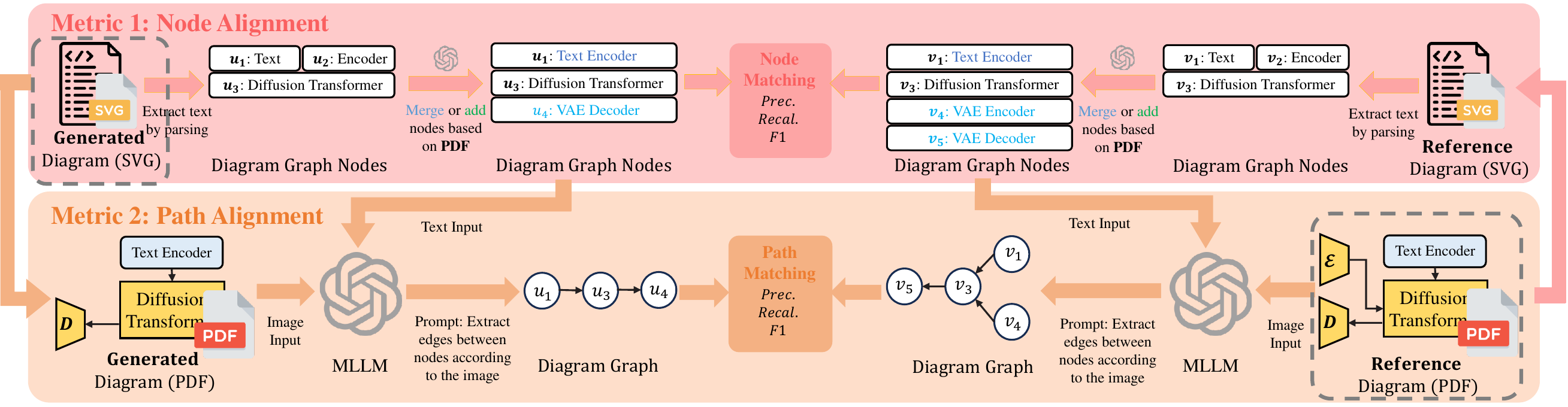}
  \caption{The detailed pipeline of \textbf{DiagramEval framework}. Intuitively, \textcolor{red}{Node Alignment} measures the correctly matched text elements between generated and groundtruth diagrams while \textcolor{orange}{Path Alignment} measures the correctly matched connections upon matched elements.}
  \label{fig:diagram_overview}
\end{figure*}
As illustrated in Figure~\ref{fig:diagram_overview}, our evaluation framework for LLM-generated diagrams is based on the core idea of treating diagrams as text-attributed graphs (TAGs)~\citep{yang2021graphformers}. In these graphs, diagram elements associated with text are defined as nodes, while connections between elements form directed edges. We introduce novel metrics to evaluate these diagrams from two complementary perspectives: \textbf{Node Alignment}, which assesses the matching of nodes between the generated and reference diagrams, and \textbf{Path Alignment}, which evaluates the consistency of paths in both diagrams connecting matched nodes.

The primary challenge in performing this detailed evaluation lies in effectively extracting nodes and edges from SVG diagram files, which we address in the following section.

\subsection{Constructing Diagram Graphs}

\paragraph{Node Extraction:} Text exists in SVG files in the form of SVG text items. Since they are accessible by scanning SVG files, the key problem of node extraction is to determine what text items belong to one node. We combine the spatial and semantic factors in a two-step process to tackle this process. 

The upper half of Figure~\ref{fig:diagram_overview} outlines the node extraction process. First, we form a draft node list by parsing the SVG file. We collect spatial coordinates and span lengths of all text items in the file. Then, we investigate every item pair about whether 1) their $y$-coordinates differ by less than $K\times$ font size, and 2) their spans in the $x$-coordinate overlap for more than $\tau$. If both conditions are fulfilled, two items are considered as one node. We apply above parsing to get the draft node list.

Second, we exploit a light-weight multi-modal LLM to refine the draft node list according to the text semantic. This step takes the rendered SVG image and draft node list as input. The LLM is tasked with several refinement operations: (1) \textbf{Merging} spatially contiguous and semantically related text nodes; (2) \textbf{Adding} missing conceptual nodes, which may include non-textual elements (such as icons or logos); and (3) \textbf{Removing} nodes with unclear semantics. We obtain the final node list $V=\{v_i\}$ from the LLM output.

\paragraph{Edge Extraction:} Edges are represented implicitly in the diagram, making their identification challenging. Manually defining rules to capture diverse visual forms of connections (e.g., arrows with various styles, lines, and spatial arrangements indicating logical flows) from raw SVG data is highly complex. Thus, we again employ an LLM with vision capabilities. As depicted in the lower half of Figure~\ref{fig:diagram_overview}, this model is provided with the rendered diagram image and a curated list of nodes, each with a unique identifier and textual content. The LLM analyzes visual cues—such as arrows, lines, and proximities—to identify all directed connections between node identifiers. This yields a set of directed edges $E = {(v_i, v_j), \dots}$, where $v_i, v_j \in V$. Combining edge set $E$ with node set $V$ results in the complete extracted graph $\mathcal{G}(V,E)$.
\\

This node and edge extraction process is independently applied to both the LLM-generated diagram $\mathcal{D}$, resulting in the graph $\mathcal{G}_{gen}$, and the groundtruth diagram $\mathcal{D}{gt}$, yielding the reference graph $\mathcal{G}_{ref}$. We evaluate its accuracy in Section~\ref{sec:exp4}.

\begin{table*}[t]
  \centering
  \setlength{\tabcolsep}{7pt} 
  \begin{tabular}{lcccccccc}
    \hline
    \multirow{2}{*}{Model}           & \multicolumn{3}{c}{Node Alignment} &\multicolumn{3}{c}{Path Alignment} & \multicolumn{2}{c}{CLIPScore}  \\
               & $prec.$ & $recal.$ & $F1$ &$prec.$ & $recal.$ & $F1$  & Text & Image \\
    \hline
    Llama 4 Maverick & \textbf{0.4737} & 0.3121 & 0.3470 & 0.2260 & 0.2506 & 0.2005 & \textbf{0.6962} & 0.6950 \\
    Gemini 2.5 Pro & 0.3600 & 0.3741 & 0.3341 & 0.2503 & \textbf{0.2817} & 0.2261 & 0.6090 & 0.7021 \\
    Claude 3.7 Sonnet & 0.2921 & \textbf{0.5087} & \textbf{0.3500} & \textbf{0.3353} & 0.2108 & \textbf{0.2419} & 0.6206 & \textbf{0.7324} \\
    \hline
  \end{tabular}
  \caption{Results on diagram generation of three LLMs over our metrics and two CLIPScore metrics~\citep{hessel2021clipscore}, the most widely-used evaluation measurement. Half or more nodes and edges in the reference diagrams are missing or not recognizable, indicating the bad quality of LLM-generated paper diagrams.
  }
  \label{tab:res}
\end{table*}

\subsection{Evaluation Metrics}

\paragraph{Node Alignment} These metrics appraise the fidelity of the textual content within the generated diagram. This involves assessing the degree to which the set of text elements $V_{gen}$ in the generated diagram aligns with the set $V_{ref}$ from the reference diagram. The core of this comparison is a node-matching procedure: each node in $V_{gen}$ is compared against all unmatched nodes in $V_{ref}$. A match is established if the textual similarity between a pair of nodes, surpasses a predefined threshold. Denoting $M_{V}$ as the set of successfully matched node pairs $(v_{gen}, v_{ref})$, where $v_{gen} \in V_{gen}$ and $v_{ref} \in V_{ref}$, we quantify performance as follows:

\begin{itemize}
\item True Positives (TP$_V$): The cardinality of the set of matched node pairs, $|M_V|$.    

\item False Positives (FP$_V$): The count of nodes in $V_{gen}$ that remain unmatched, $|V_{gen}| - \text{TP}_V$.

\item False Negatives (FN$_V$): The count of nodes in $V_{ref}$ that remain unmatched, $|V_{ref}| - \text{TP}_V$.

\end{itemize}

Based on these quantities, we compute information retrieval metrics—Precision$_V$, Recall$_V$, and F1-score$_V$ as our proposed metrics for node alignment.

\paragraph{Path Alignment} We also assess the node-wise structural fidelity encoded in $\mathcal{G}_{gen}$ corresponds with that in $\mathcal{G}_{ref}$ in addition to node alignment. We choose to investigate \textit{path}, whose existence is a reachability indicator between two nodes. Not limited to neighborhood, paths reveal all relationships in the diagram, thus being a better feature for relational information comparison. 

Crucially, this comparison is constrained to the subgraph induced by the previously matched nodes $M_V$, because node appearance has been evaluated in the Node Alignment. Specifically, we exclude those unmatchable nodes in one graph that are not possible to get involved a path in the other graph.

Let $M_V = {(v_{gen}, v_{ref})}$ denote the set of matched node pairs. We define $V_M$ as the set of matched nodes, i.e., $V_M = \{v_{gen}\ |\ (v_{gen}, v_{ref}) \in M_V\} = \{v_{ref}\ |\ (v_{gen}, v_{ref}) \in M_V\}$. For simplicity, we maintain a one-to-one correspondence between nodes in $V_{gen}^M$ and $V_{ref}^M$ via the mapping defined by $M_V$.

We then induce subgraphs $\mathcal{G}_{gen}^{M}$ and $\mathcal{G}_{ref}^{M}$ on the matched nodes $V_M$ in both the generated and reference graphs, respectively.

For each ordered pair of distinct matched nodes $(u, v)$ where $u, v \in V_M$ and $u \neq v$, we assess: 1) whether there exists a path from $u$ to $v$ in $\mathcal{G}_{gen}^{M}$, and 2) whether there exists a path from the corresponding node $u'$ to $v'$ in $\mathcal{G}_{ref}^{M}$, where $(u, u') \in M_V$ and $(v, v') \in M_V$. Formally, we define:
\begin{equation}
\begin{aligned}
P_{gen} =& \{(u, v)\ |\ u \neq v, \\
&\text{ path from } u \text{ to } v \text{ exists in } \mathcal{G}_{gen}^{M}\}\\
P_{ref} =& \{(u, v)\ |\ u \neq v, \\
&\text{ path from } u' \text{ to } v' \text{ exists in } \mathcal{G}_{ref}^{M}\}
\end{aligned}
\end{equation}
where, for each $(u, v) \in P_{gen}$ or $P_{ref}$, $u$ and $v$ are matched nodes and $u'$ and $v'$ are their respective counterparts in the other graph according to $M_V$. We can then compare $P_{gen}$ and $P_{ref}$ by defining:

\begin{itemize}
\item {True Positives (TP$_P$)}: The number of node pairs for which a path exists in both induced subgraphs, i.e., $|P_{gen} \cap P_{ref}|$.
\item {False Positives (FP$_P$)}: Node pairs where a path exists only in $\mathcal{G}_{gen}^{M}$, i.e., $|P_{gen} \setminus P_{ref}|$.
\item {False Negatives (FN$_P$)}: Node pairs where a path exists only in $\mathcal{G}_{ref}^{M}$, i.e., $|P_{ref} \setminus P_{gen}|$.
\end{itemize}

Based on these quantities, we compute Precision$_P$, Recall$_P$, and F1-score$_P$ as our metrics for path alignment.
\\

In concert, metrics for \textbf{Node Alignment} and \textbf{Path Alignment} furnish a fine-grained and explainable evaluation of LLM-generated diagrams. They discriminate diagrams by exactly telling the mismatching of text elements and their relationship. Our advantage in interpretablity also provides guidance on where the generation could be improved. In the next section, we support this point by results.

\section{Experiment}
\label{sec:exp}

This section discuss our experiments conducted to validate our metrics by comparison with CLIPScore~\citep{hessel2021clipscore}, the most widely-used evaluation metrics for diagram generation. We first explain our experiment setup in Section~\ref{sec:exp1}. Then, we give the main result of our experiment in Section~\ref{sec:exp2}. Section~\ref{sec:exp3} demonstrates the statistics of metrics in the experiment. Section~\ref{sec:exp4} validates our metrics by comparing to human evaluation. Section~\ref{sec:exp5} explains with cases what happen when our metrics and CLIPScore differ, respectively. Throughout our experiment, we show that our metrics provide unique and beneficial information towards better evaluation of automated diagram generation.

\subsection{Experimental Setup}
\label{sec:exp1}
As mentioned in Section~\ref{sec:bg}, we first prompt state-of-the-art LLMs to generate diagrams based on the text input. Then, we evaluate the generated diagrams over our 6 metrics ($3\times$ Node Alignment, $3\times$ Path Alignment) and CLIPScore~\citep{hessel2021clipscore}, the most common metric for diagram generation. Following are our detailed setup:

\paragraph{Diagram Generation} We pick three cutting-edge LLMs for diagram generation: Llama 4 Maverick, Gemini 2.5 Pro, and Claude 3.7 Sonnet, which we access by their official APIs. The unified prompts we use to generate diagrams are omitted to Appendix~\ref{appendix:1}. The layout caption is generated by Gemini-2.0-Flash-lite by prompting to generate a layout caption for the diagram image. The used prompts are omitted to Appendix~\ref{appendix:1}.

\paragraph{DiagramEval} We use Gemini-2.0-Flash-lite as the LLM used in our evaluation pipeline, including node extraction refinement and edge extraction. The used prompts are omitted to Appendix~\ref{appendix:1}. $K$ and $\tau$ in the node extraction are empirically set to $1.5$ and $0.2$. 

\paragraph{Baseline} We use the code~\footnote{\url{https://github.com/aszala/DiagrammerGPT}} of DiagrammerGPT~\citep{zala2023diagrammergpt} for computing CLIPScore. Following their implementation, we use SigLIP~\citep{zhai2023sigmoid}~\footnote{\url{https://huggingface.co/google/siglip-so400m-patch14-384}} as the vision and language encoder in CLIPScore. LLM-generated layout captions and groundtruth diagrams are selected as text reference and image reference for CLIPScore, respectively. Following DiagrammerGPT~\citep{zala2023diagrammergpt}, we use model-generated captions for computing CLIPScore because original captions may not cover enough details of the groundtruth diagram.
\begin{figure*}[htbp]
\vspace{-0.5cm}
\centering
  \includegraphics[width=0.9\linewidth]{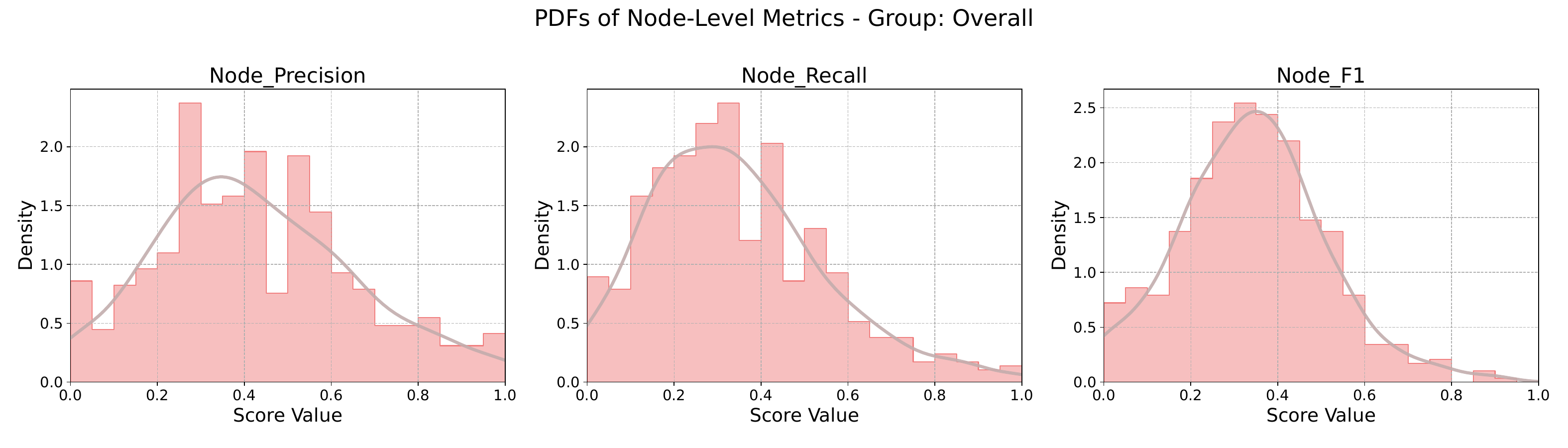}
  \includegraphics[width=0.9\linewidth]{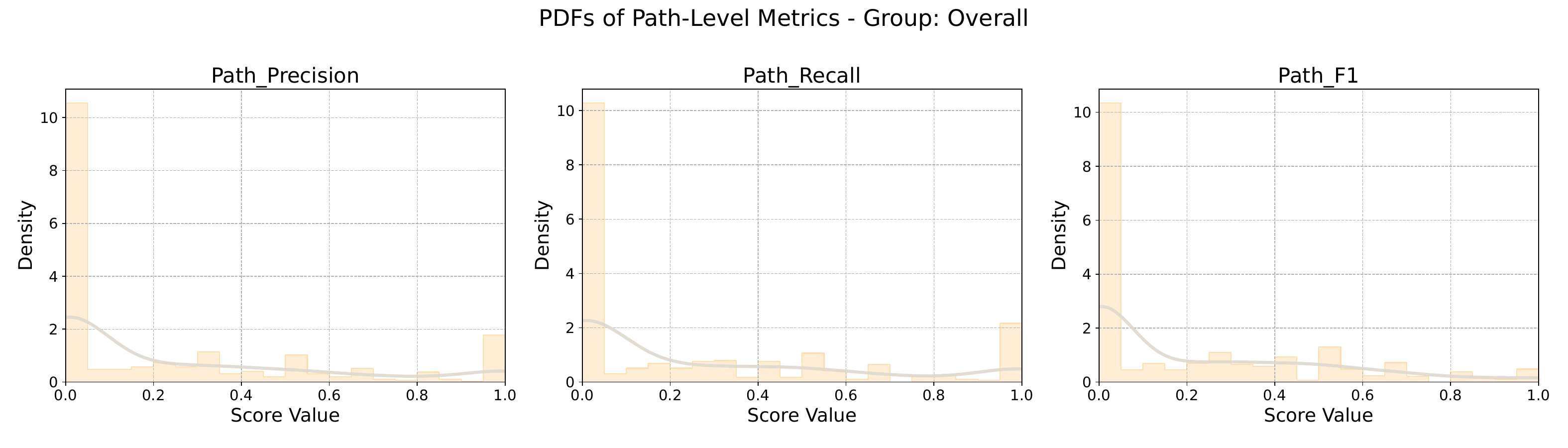}
  \includegraphics[width=0.6\linewidth]{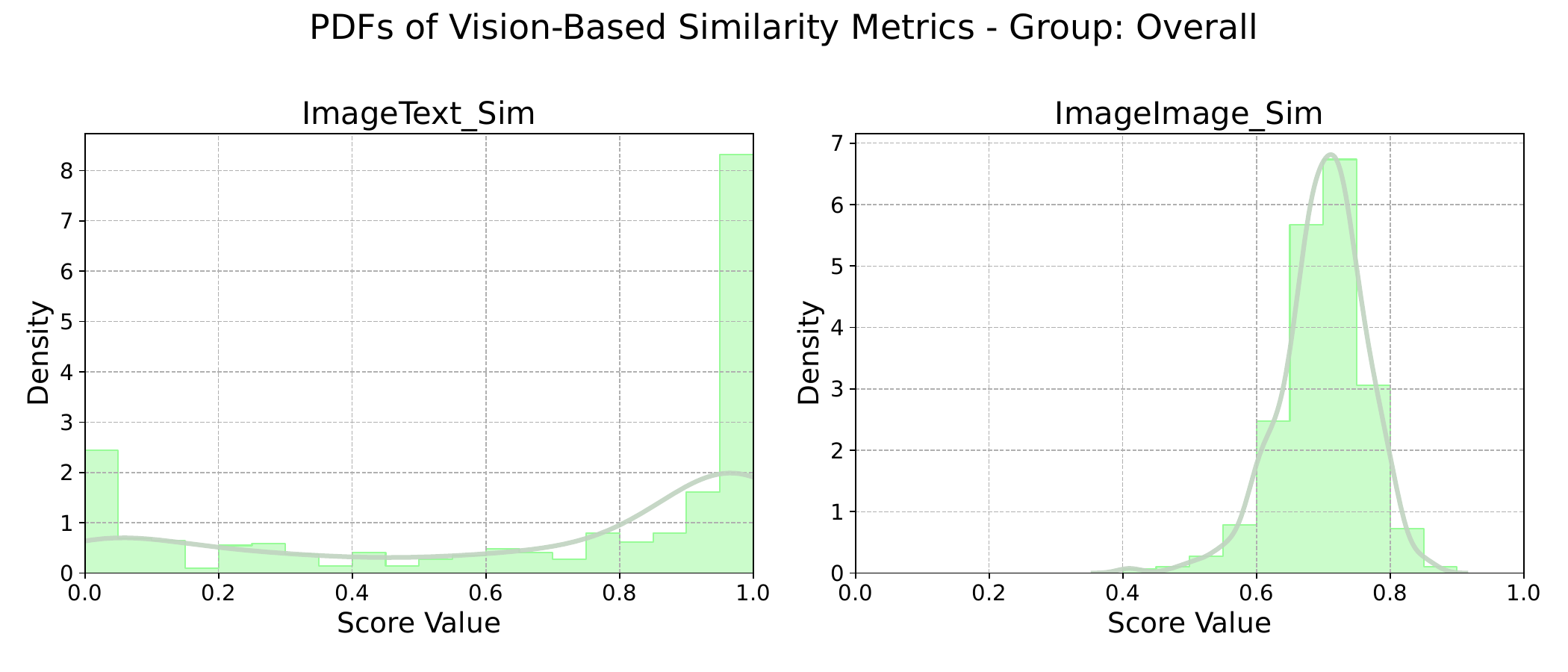}
  \caption{Statistic results: probability density functions (PDFs) of our 6 novel metrics and 2 CLIPScore.}
  \label{fig:stats1}
  \vspace{-0.5cm}
\end{figure*}
Notably, VPEval~\citep{cho2023visual} is not suitable for evaluating paper diagrams. First, the object detection in VPEval needs text to explicitly provide the objects. However, in paper diagrams, there are often dozens of objects while few of them are explicitly described by paper context. Hence, we cannot use VPEval to evaluate paper diagrams. Second, VPEval only counts direct connections between objects. However, generated diagrams tend to skip some connections in the reference diagrams, for example, a-c compared to the original a-b-c. This is normal in most cases when b is not very important compared to the correct connection between a and c, for example, b is a linear layer between two backbone models. However, VPEval cannot capture such indirect connections, thus losing discrimination.

\paragraph{Dataset} To avoid knowledge leakage, we collect papers accepted by CVPR2025 as the source of our evaluation dataset, because they are released after the data cutoff date of three LLMs. We use an automated pipeline to select diagrams with abundant text annotation and collect them with captions and corresponding paper context. A total number of 361 items are included in our dataset. Data consensus and licenses are omitted to Appendix~\ref{appendix:2}.

\subsection{Quantitative Result}
\label{sec:exp2}

Table~\ref{tab:res} shows our quantitative result. Three models have similar performance in diagram generation over our metrics and CLIPScore, proving our basic soundness. All metrics agree that Claude generates diagrams that best align with the groundtruth diagrams, for it performs the best over 4 out of our 6 metrics and CLIPScore (Image). 

One interesting observation is that Claude suffers from poor node alignment precision while having the outstanding node recall. One potential explanation is that its generated diagrams tend to include extraneous nodes compared to groundtruth diagrams. To validate this assumption, we count the average number of nodes in generated diagrams by three LLMs. The result fits our assumption that Claude produces diagrams with 31.67 nodes on average, much more than 21.42 nodes of Gemini and 10.68 nodes of Llama. This may also explain the poor performance of Claude over CLIPScore (Text): While CLIPScore (Image) puts weights on the layout and visual elements, CLIPScore (Text) is more sensitive to unexpected text. Hence, irrelevant text elements greatly affect Claude's performance over CLIPScore (Text). 

This observation highlights how our metrics provide interpretable insights into diagram generation performance, complementing coarse-grained metrics with fine-grained, structure-aware evaluation.

\subsection{Statistics}
\label{sec:exp3}

\begin{figure}[htbp]
\centering
  \includegraphics[width=0.8\linewidth]{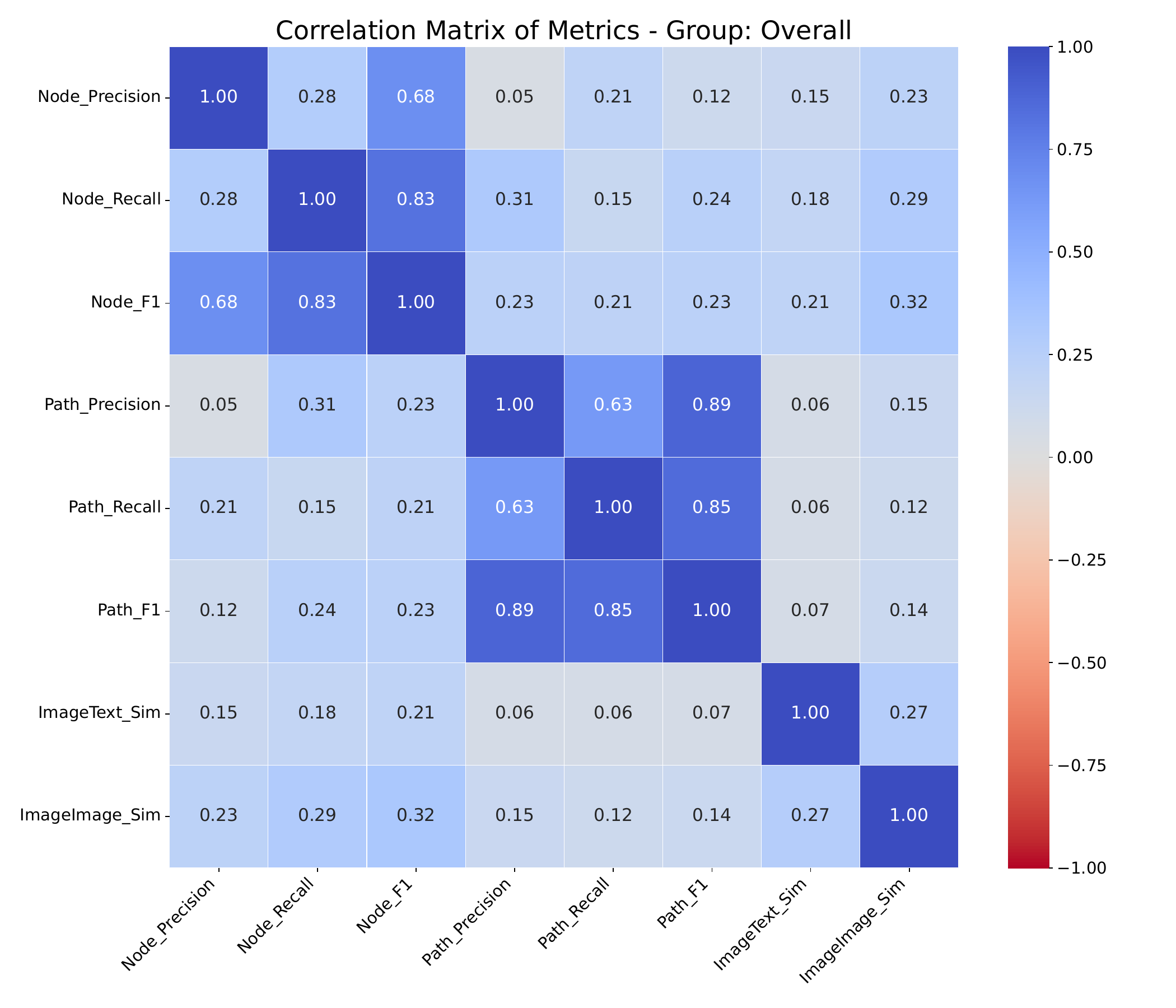}
  \caption{Correlation map of our 6 novel metrics and 2 CLIPScore metrics. Metrics of Node Alignment show considerable positive correlation with 2 CLIPScore metrics. Metrics of Path Alignment appear to be indifferent with 2 CLIPScore metrics.}
  \label{fig:stats2}
\end{figure}

We compute the probability density functions of our 6 novel metrics and 2 CLIPScore~\citep{hessel2021clipscore}, whose result is given in Figure~\ref{fig:stats1}. We also analyze their correlation and demonstrate the result in Figure~\ref{fig:stats2}.

Node Alignment metrics showed a notable positive correlation with CLIPScore metrics, likely due to their common focus on text element consistency. However, Node Alignment exhibited healthier score distributions. This improvement stems from isolating node-level textual alignment from factors like spatial layouts, colors, and styles, which significantly affect CLIPScore despite being irrelevant to textual content accuracy. Simply changing the layout from vertical to horizontal can partially offset the drop in CLIPScore caused by removing all connections in the diagram. Additionally, CLIPScore's sensitivity to superficial image elements also inherently limits scoring extremes, constraining its effectiveness in diagram evaluations. Given the fact that novel methods often report only $0.01$ improvements on CLIPScore~\citep{zala2023diagrammergpt,mondal2024scidoc2diagrammer}, \textbf{Node Alignment is a good complement and a potential alternate to CLIPScore}.

Conversely, Path Alignment metrics displayed minimal correlation with CLIPScore. The case study in the next section will show that this is because many diagrams generated by three LLMs, even they are state-of-the-art ones, perform poorly in expressing relationship. This finding is co-validated by existing research on whether LLM can understand graph structures within text~\citep{wang2023can}. However, this shortcoming has never been explicitly revealed by any existing metrics. \textbf{Path Alignment} provides fine-grained, interpretable insights into missing or incorrect connections, \textbf{being a novel perspective of evaluation not offered by existing metrics.}
\begin{figure*}[htbp]
\centering
  \includegraphics[width=\linewidth]{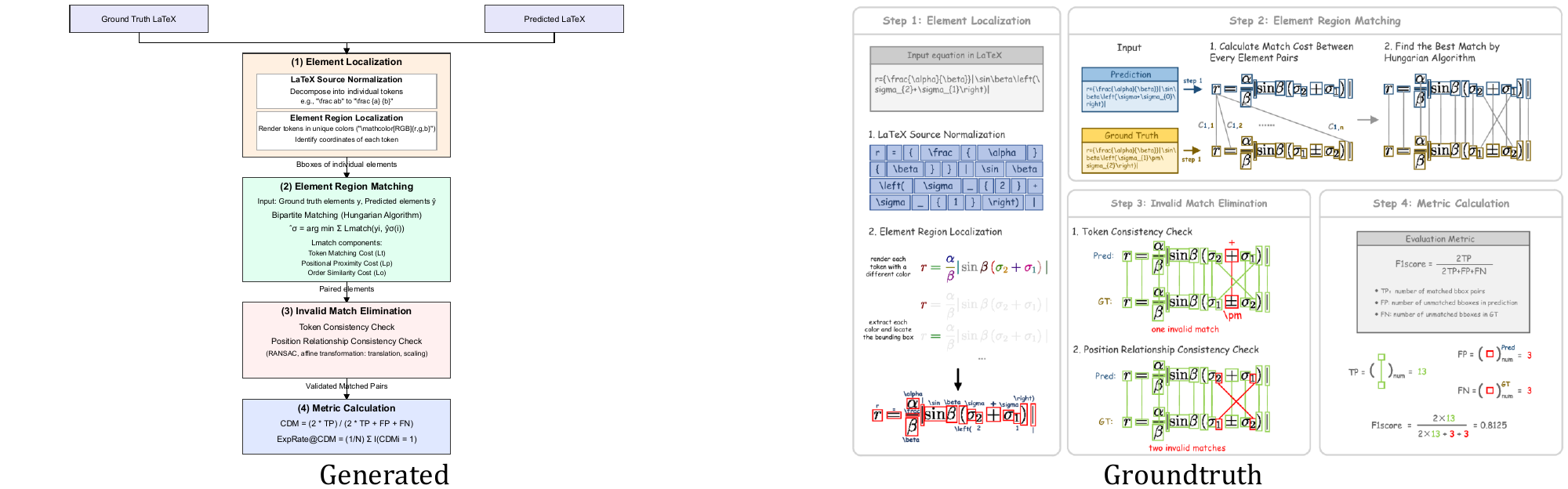}
  \caption{Case: Low CLIPScore (Text) and high Path F1. \textbf{CLIPScore (Text):} 0.2558. \textbf{Path F1:} 1.}
  \vspace{5mm}
  \label{fig:case1}
  \includegraphics[width=\linewidth]{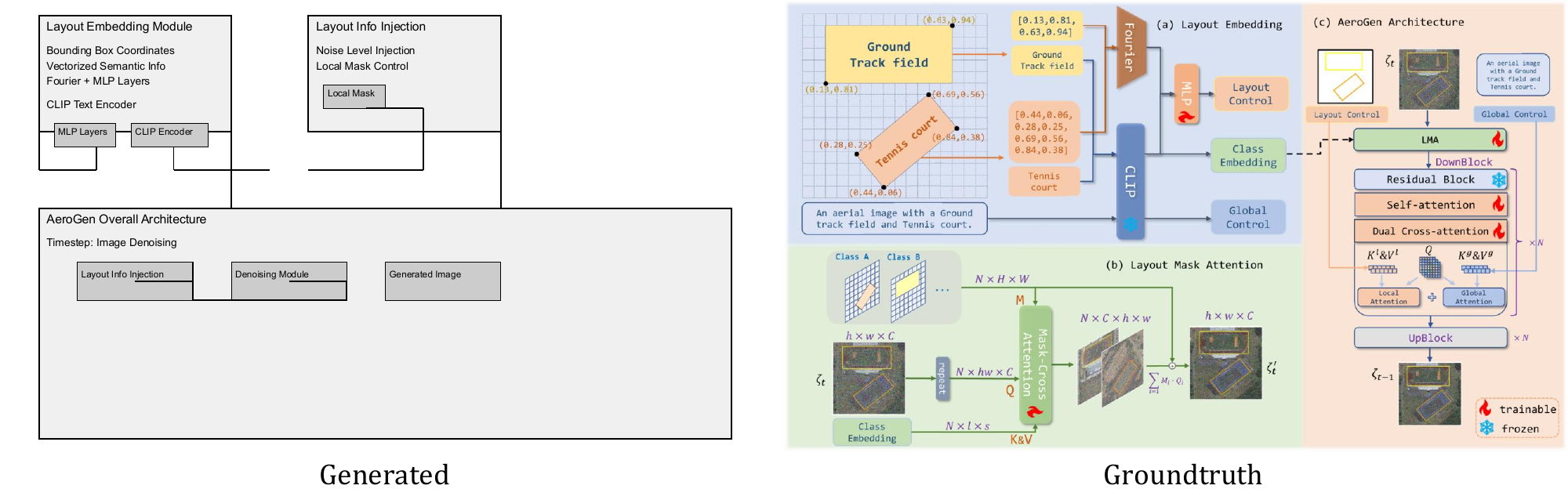}
  \caption{Case: High CLIPScore (Text) and low Path F1. \textbf{CLIPScore (Text):} 1. \textbf{Path F1:} 0.}
  \label{fig:case2}
\end{figure*}
\subsection{Human Evaluation}
\label{sec:exp4}
\textbf{Accuracy of Node Extraction:} Our node extraction simply transfers one single text element in the SVG format into one node. This is accurate because phrases are naturally placed in one text element in both reference and generated diagrams. 2) Edge extraction: We conduct new experiments to examine the accuracy of edge extraction. We randomly pick 1 edge for every diagram and let two machine learning researchers judge whether this edge exists in the diagram. 

The result shows that 85.87\% of the nodes in reference diagrams and 90\% in generated diagrams are extracted accurately, where 361/361 of reference diagrams and 300/361 of generated diagrams are evaluated (no edge cannot be extracted from the rest 61 generated diagrams because of the low quality). While it is hard to know how many edges are there in the diagram, these precision scores make sure that extracted edges are highly possible to exist in the diagram. This validates the accuracy of our edge extraction process. We will add this result to our new draft to complement the paper.

\textbf{Correlation with human evaluation:} We follow the human evaluation of \citet{cho2023visual} and select a subset of 50 reference diagrams (the first 50 by the name order) and corresponding generated diagrams by Gemini-2.5-Pro. Two senior machine learning researchers then evaluate the semantic similarity between reference and generated diagrams by answering the question: do two diagrams express the same logic? Our interface offers three options for the human evaluators: good (1.0), fair (0.5), and bad (0). The average similarity score is 0.3298, with nearly half of the results being 0. The following Table~\ref{tab:correlation_metrics} shows the correlation between the human-evaluated similarity score and metrics in our papers:
\begin{table}[h]
\centering
\begin{tabular}{lr}
\hline
\textbf{Metric} & \textbf{Correlation} \\ \hline
Node F1 & 0.4316 \\ 
Path F1 & 0.4052 \\ 
CLIPScore-Text & 0.1065 \\ 
CLIPScore-Image & 0.0831 \\ \hline
\end{tabular}
\caption{Correlation with human evaluation scores of different metrics. Our metrics align better with the judgment of human experts.}
\label{tab:correlation_metrics}
\end{table}
\begin{figure*}[htbp]
\centering
  \includegraphics[width=\linewidth]{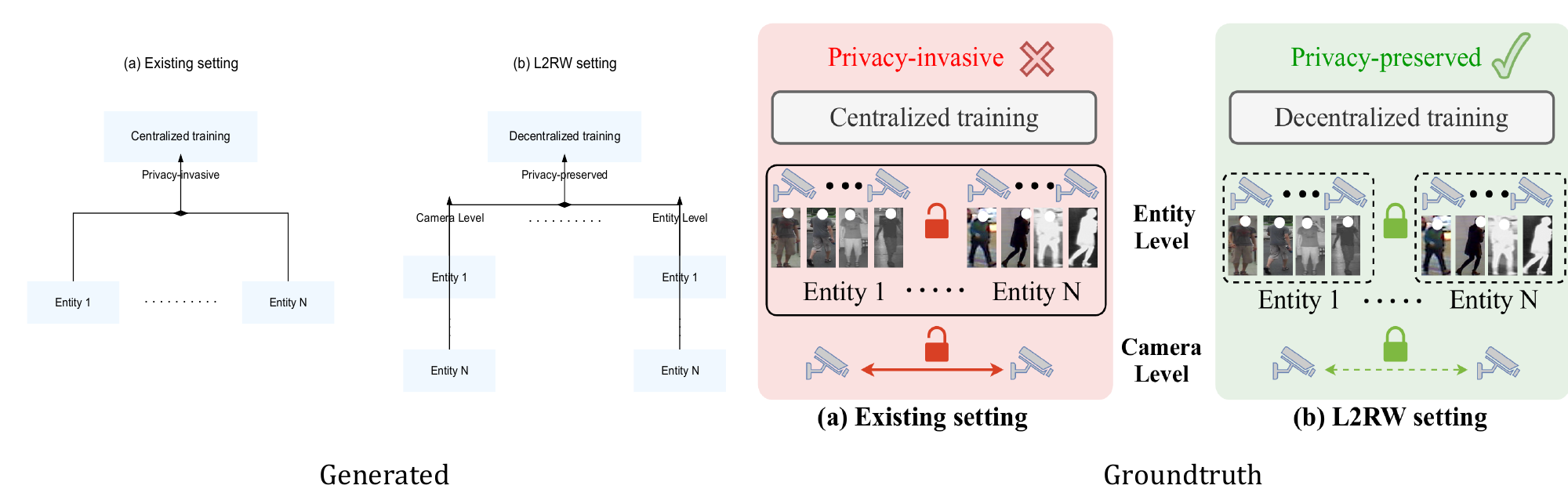}
  \caption{Case: Low CLIPScore (Image) and high Node F1. \textbf{CLIPScore (Image):} 0.6007. \textbf{Node F1:} 0.8696}
  \label{fig:case3}
  \vspace{5mm}
  \includegraphics[width=\linewidth]{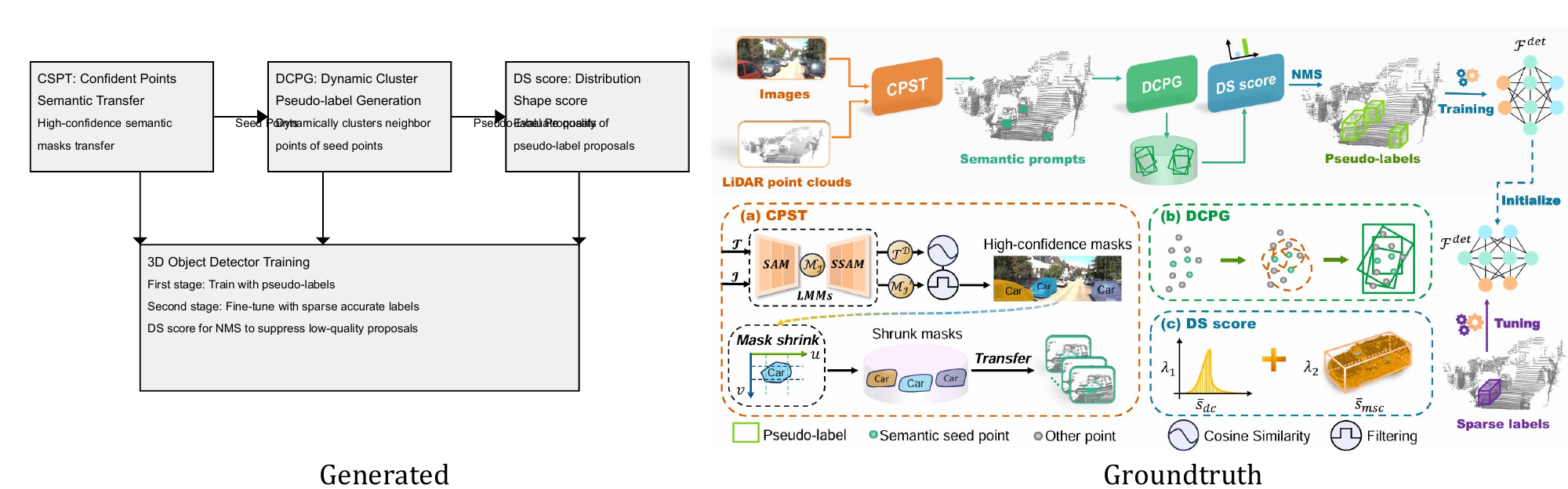}
  \caption{Case: High CLIPScore (Image) and low Node F1. \textbf{CLIPScore (Image):} 0.8094. \textbf{Node F1:} 0.16}
  \label{fig:case4}
\end{figure*}
Our new metrics show better alignment with human evaluation. We observe that CLIPScore-Image may give a generated diagram a relatively good score even if its element-wise logic is definitely different from the reference, while CLIPScore-Text is too sensitive to the text description. While our metrics are more aligned to human judgment compared to CLIPScore, the most widely-used metric in the field, we believe they are trustworthy enough to provide another perspective of evaluation to generated diagrams.

\subsection{Case Study: When and how our metrics and CLIPScore differ}
\label{sec:exp5}
The metric statistics show that our new metrics does not fully consistent with CLIPScore, the classical evaluation metric. This raises a research question: What happen when our metrics and CLIPScore differ? In this section, we select four cases with distinct scores by our metrics and CLIPScore and explain why the difference takes place.

Figure~\ref{fig:case1} shows a case with low CLIPScore (Text) and high Path F1. We can see an explicit data flow in the generated diagram, with good alignment with the semantic of groundtruth diagram as well as the original paper~\citep{wang2024cdm}. This is well captured by our Path F1. However, the spatial layout and icons in the generated diagram are very different from those in the groundtruth. While layout captions used in CLIPScore (Text) include detailed description to these visual elements, the generated diagram performs poorly over CLIPScore (Text). This case consolidates the point that the sensitivity to visual elements of CLIPScore hinder its recognition of good diagrams. More robust to the interference of layouts and icons, our metrics complement this disadvantage.

Figure~\ref{fig:case2} shows a case with high CLIPScore (Text) and low Path F1. In contrast to the above case, there is no clear data flow. Hence, our Path F1 assigns 0 to this diagram. However, the generated diagram includes all text mentioned in the layout caption. For this reason, it has perfect alignment with the layout caption under CLIPScore (Text). This exposes an inherent problem of depending evaluation on models: the generator may hack the metric. Only by placing all text elements mentioned in the layout caption (rather than those existing in the original diagram), the generated diagram yields perfect score in the model-based comparison with the layout caption. Our metrics ease this problem by obtaining intermediates from models.

Figure~\ref{fig:case3} shows a case with low CLIPScore (Image) and high Node F1. Similar to the case in Figure~\ref{fig:case1}, the evaluation of CLIPScore is interfered by the difference in spatial layout and non-textual icons. By contrast, our Node F1 focuses only on the existence of text elements, thus giving a more objective result. Figure~\ref{fig:case4} suffers similar problems, where CLIPScore (Image) ignores detailed logic and evaluates the diagram unreasonably. 

To conclude, our metrics overcome two shortcomings of CLIPScore: \textbf{visual element interference} and \textbf{metric hacking} in these four cases. This proves that our metrics constitute good complements to CLIPScore.

\section{Conclusion}
This paper introduces DiagramEval, a novel set of metrics for evaluating large language model (LLM)-generated scientific diagrams. Unlike existing evaluation methods, DiagramEval represents diagrams as graphs and performs automated extraction of graph structures from diagrams. Evaluation metrics are then computed through fine-grained comparisons between the nodes and paths in the generated and reference graphs. DiagramEval addresses the current lack of fine-grained, interpretable, and structure-aware metrics in the assessment of automated diagram generation.

\section*{Limitations} Our proposed metrics have limitations primarily due to uncertainties in LLM performance. Specifically, the LLM may not reliably identify all edges in the reference graph, potentially causing inaccurate or underestimated evaluations. Despite this, our metrics reduce dependency on complex models and offer more interpretable outcomes compared to existing methods. Addressing these limitations is an important direction for future work.

As our work focuses on evaluating diagram generation, it does not raise new potential risks other than those general ones by using LLMs to generate contents.

\section*{Acknowledgments}
We sincerely appreciate the support from Amazon grant
funding project \#120359, "GRAG: Enhance RAG Applications with Graph-structured Knowledge", and Meta gift
funding project "PERM: Toward Parameter Efficient Foundation Models for Recommenders".

\bibliography{custom}

\appendix

\section{Appendix}
\label{sec:appendix}

\subsection{Data License and Consensus} 
\label{appendix:2}
In this study, we utilize a collection of research papers from arXiv as our primary data source. To ensure ethical and legal reuse, we include only papers published under open-access licenses that permit redistribution, including Creative Commons Attribution (CC BY 4.0), Attribution-ShareAlike (CC BY-SA 4.0), and Attribution-NonCommercial-ShareAlike (CC BY-NC-SA 4.0). While the ShareAlike and NonCommercial terms impose certain restrictions—such as requiring derivative works to be shared under the same license or prohibiting commercial use—we fully comply with these conditions, using the materials only for academic, non-commercial research with appropriate attribution. Due to the fact that we only conduct text mining on the papers and that the number of papers is huge, we do not cite these papers.
\subsection{Usage of AI Assistant}
We use ChatGPT to polish the text of our paper.


\subsection{Prompts}
\label{appendix:1}
 Following are four prompts we use in our experiment for 1) layout caption generation, 2) diagram generation, 3) node extraction refinement, and 4) edge extraction, respectively.

\begin{lstlisting}[language=Python, caption=Prompt: 
Layout Caption Generation]
prompt_parts = [
            "Describe the spatial layout of the components in this document, focusing on their relative positions and connections.",
            "For example: 'Component A is above Component B, and an arrow connects B to C which is to the right of A'.",
            "Do not interpret the meaning of the diagram, only its visual structure and element arrangement. Be concise.",
            pdf_blob_part
        ]
\end{lstlisting}

\begin{lstlisting}[language=Python, caption=Prompt: Diagram Generation]
prompt = f"""
<INSTRUCTION>

Generate an SVG diagram based on the following information.

**Rules:**
1.  Create clean, well-structured SVG code. Keep the diagram width="1000" height="700".
2.  Use main concepts and expressions given in the original paper context for element text (very important).
3.  Ensure elements (shapes, text) do not overlap.
4.  Do **not** include any legends.
5.  Arrows must start and end precisely on the border of the elements they connect. Arrows should avoid crossing other elements by using vertical and horizontal corner arrows. Do not use any sloping arrows.
6.  Represent the core mechanisms described in the context. Avoid using a single large block for a complex mechanism that should be broken down. But also keep the mechanism representation intuitive and easy-to-understand enough.
7.  **Never** use any characters leading to SVG rendering issues, for example, & (Ampersand).
8.  Keep proper layout tightness. Don't leave a lot meaningless blank space between elements.
9.  Add font-size independently to every single text element.
10. Avoid generating problematic svg code, for example, svg code with invalid xml characters or duplicate attributes.
{'11.  Adhere strictly to the spatial layout in the layout and element text.' if spatial_layout_prompt else ''}

Please output *only* the SVG code block, starting with `<svg` and ending with `</svg>`.

<END OF INSTRUCTION>

**Paper Context:**
{paper_context}

**Diagram Caption/Focus:**
{diagram_caption}
{layout_section}

Now, output the SVG code block:
"""
\end{lstlisting}

\begin{lstlisting}[language=Python, caption=Prompt: 
Node Extraction Refinement]
prompt_parts = [
            "You are an expert diagram analysis assistant specializing in text element coherence.",
            f"The following image is a '{diagram_type_name}'. I have already performed an initial text extraction from its source, resulting in the list of text elements below.",
            "\n**Image of the diagram:**",
            pil_image,
            "\n\n**Currently Extracted Textual Elements (- Element [ID]: \"[TEXT]\"):**",
            element_list_str,
            "\n\n**Your Task:**",
            "Analyze the image and the provided list of elements. Your goal is to improve the element list by identifying necessary merges, additions, or removals.",
            "\n1. **Merges**: Identify if any listed elements are parts of a single, continuous text block in the image and should be merged.",
            "   For example, if 'Element ID_A: Hello' and 'Element ID_B: World' visually form 'Hello World', they should be merged.",
            "\n2. **Additions**: Identify two specific types of missing nodes:",
            "   a) Duplicate nodes: Nodes that have the same text as existing nodes but represent different instances in the diagram.",
            "      For example, if there are two 'mask' nodes in the diagram but only one is in the current list.",
            "   b) Non-text nodes: Nodes that use icons or images instead of text to represent concepts.",
            "      For example, an OpenAI logo representing LLMs, or a neural network icon representing a model.",
            "      For these nodes, generate appropriate text descriptions based on their visual representation.",
            "\n3. **Removals**: Identify nodes that should be removed based on the following strict policies:",
            "   a) Non-English/Non-Math Text: Remove nodes that contain **ONLY** non-English and non-mathematical characters.",
            "      For example, if a node contains **only** Chinese characters, it should be removed.",
            "      However, if the node contains a mix of English and non-English text, keep it.",
            "   b) Numbers Only: Remove nodes that contain **ONLY** numbers (including decimal points and basic math symbols).",
            "      For example, '123', '3.14', or '1+2' should be removed.",
            "   c) Non-conceptual elements: Remove nodes not representing concepts in the diagram, such as text explanation, description, or examples."
            "\n**Important Notes:**",
            "- Do not consider general diagram elements (like arrows, lines, or decorative elements) as nodes to be added.",
            "- For duplicate nodes, ensure they are truly separate instances in the diagram.",
            "- For non-text nodes, generate clear and concise descriptions that capture their meaning.",
            "- For removals, strictly follow the three policies above. **Do not remove nodes for any other reasons.** ",
            "\n**Output Format:**",
            "First, provide your analysis in a 'Thinking Phase' section, explaining your observations and reasoning.",
            "Then, after the signal 'FINAL ANSWER:', provide your findings as a JSON object with three optional keys: 'merges', 'adds', and 'removes'.",
            "- 'merges': A list of objects, where each object has 'keep_id' (the ID of the element to retain and append to) and 'remove_id' (the ID of the element whose text will be appended and then the element removed).",
            "- 'adds': A list of objects, where each object has a 'text' key for the newly identified text string.",
            "- 'removes': A list of objects, where each object has a 'id' key for the element ID to be removed.",
            "Example JSON output: {\"merges\": [{\"keep_id\": \"G_1\", \"remove_id\": \"G_2\"}], \"adds\": [{\"text\": \"LLM Model (OpenAI)\"}, {\"text\": \"Input Image\"}], \"removes\": [{\"id\": \"G_3\"}, {\"id\": \"G_4\"}]}.",
            "If no operations are needed, provide an empty JSON object {} or omit keys.",
            "Only include IDs from the provided list for merging and removing. Ensure 'keep_id' and 'remove_id' are different.",
            "\n**Response Structure:**",
            "1. Start with 'THINKING PHASE:' and provide your detailed analysis",
            "2. After your analysis, write 'FINAL ANSWER:' on a new line",
            "3. Then provide the JSON output"
        ]

\end{lstlisting}

\begin{lstlisting}[language=Python, caption=Prompt: 
Edge Extraction]
prompt_parts = [
            "You are an expert diagram analysis assistant.",
            f"The following image is a diagram ({diagram_type}). I have already extracted the text elements from it.",
            "\n\n**Identified Textual Elements in the {diagram_type} Diagram (- Element [ID]: \"[TEXT]\"):**",
            element_list_str,
            "\n\n**Task:**",
            f"Analyze **all** connections (e.g., arrows, lines indicating flow) in the {diagram_type} Diagram image.",
            "Identify **all** DIRECTED one-to-one connections BETWEEN the provided element IDs.",
            "Every element should involve in at least one connection."
            "All straight or corner arrows indicate connections",
            "First, think step-by-step about the connections. Then, on a new line, provide the final list of connections.",
            "Output your findings as a JSON list of lists, where each inner list is a pair of element IDs representing a directed connection from the first ID to the second ID.",
            "For example: [[\"ID1\", \"ID2\"], [\"ID1\", \"ID3\"], [\"ID4\", \"ID2\"]].",
            "Only include **IDs** (not the text) from the list provided above. Ensure the source and target IDs are correct based on the diagram's flow.",
            "If there are no connections, return an empty list [].",
            "Start your final JSON output with the signal 'Final Answer JSON:'.",
            "\n\n**{diagram_type} Diagram Image:**",
            pil_image,
            "\n\n**Thinking Process and JSON Output of Connections:**"
        ]
\end{lstlisting}

\end{document}